\documentclass{article}
\usepackage{arxiv}
\usepackage[T1]{fontenc}
\usepackage[utf8]{inputenc}
\usepackage{amsmath,amssymb,graphicx,booktabs}
\usepackage[numbers,sort&compress]{natbib}
\usepackage[dvipsnames]{xcolor}
\usepackage[hyphens]{url}
\usepackage{microtype}
\usepackage{placeins}
\usepackage[font=small,labelfont=bf,labelsep=period]{caption}
\definecolor{linkblue}{rgb}{0.15,0.32,0.58}
\usepackage[breaklinks,colorlinks,allcolors=linkblue]{hyperref}
\hypersetup{pdftitle={Palmprint Decomposer: Interpretable and Editable Factorization via Synthetic Learning and Instance-Wise Refinement},pdfauthor={Licheng Yan, Yuqi Wang, Jingtao Wang, Bob Zhang, Andrew Beng Jin Teoh, Lu Leng, Shuyi Li, Ziyuan Yang},pdfsubject={Palmprint image factorization},pdfkeywords={palmprint, image decomposition, synthetic learning, instance-wise refinement}}
\renewcommand{\shorttitle}{Palmprint Decomposer}
\renewcommand{\headeright}{A Preprint}
\title{Palmprint Decomposer: Interpretable and Editable Factorization via Synthetic Learning and Instance-Wise Refinement}
\author{\normalfont Licheng Yan$^{1}$, Yuqi Wang$^{1}$, Jingtao Wang$^{1}$,\\
Bob Zhang$^{1,*}$, Andrew Beng Jin Teoh$^{2}$,\\
Lu Leng$^{3}$, Shuyi Li$^{1}$, Ziyuan Yang$^{4}$\\[3pt]
{\footnotesize\normalfont $^{1}$PAMI Research Group, Department of Computer and Information Science, University of Macau, Macau, China}\\
{\footnotesize\normalfont $^{2}$School of Electrical and Electronic Engineering, College of Engineering, Yonsei University, Seoul 120749, Republic of Korea}\\
{\footnotesize\normalfont $^{3}$School of Software, Nanchang Hangkong University, Nanchang, Jiangxi 330063, P. R. China}\\
{\footnotesize\normalfont $^{4}$School of Cyber Science and Engineering, Sichuan University, Chengdu, Sichuan 610065, P. R. China}\\
{\footnotesize\normalfont $^{*}$Corresponding author: Bob Zhang (\href{mailto:bobzhang@um.edu.mo}{\texttt{bobzhang@um.edu.mo}})}
}

\date{}
\begin{document}
\maketitle
\begin{abstract}
Palmprint recognition and synthesis methods focus on extracting discriminative textures or generating images with varied textures and appearances. However, they do not generally provide an explicit image-level representation that separates palmprint texture from its underlying base appearance for independent analysis or controlled editing. We introduce Palmprint Decomposer, an interpretable framework that represents a palmprint through independently editable factors and reconstructs the image through a multiplicative composition model. A color map and a light map characterize the base appearance, while a structure map and a global texture strength parameter describe texture modulation. However, accurate reconstruction alone does not guarantee separation of palmprint texture from base appearance, as fine palmprint textures may leak into the estimated base appearance. To mitigate this ambiguity, we train the decomposer exclusively on synthetic data with direct factor supervision, using asymmetric decoding paths and decomposition constraints to suppress texture leakage. Instance-Wise Refinement further improves real-image reconstruction by directly optimizing the predicted factors for each input without relying on the trained network. Experiments across seven datasets demonstrate high reconstruction fidelity, with the extracted structure maps generally achieving verification performance comparable to the original images. Compared with existing methods, our approach achieves better factor separation while preserving more discriminative palmprint texture information.
\end{abstract}

\section{Introduction}
\label{sec:introduction}
\begin{figure}[t]
	\centering
	\includegraphics[width=0.46\linewidth]{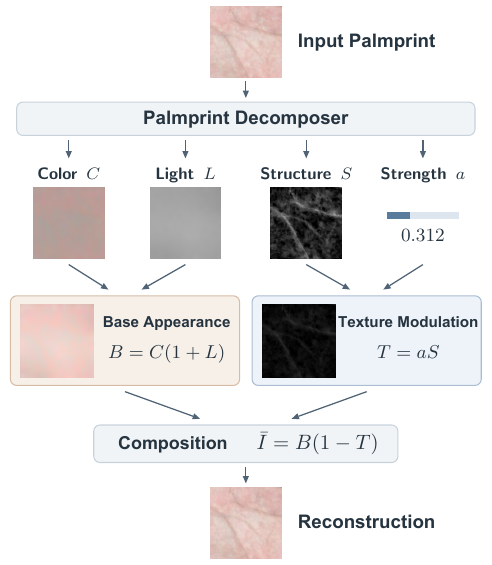}
	\caption{Palmprint decomposition and recomposition. The color map $C$ and light map $L$ define base appearance $B$, while the structure map $S$ and global texture strength $a$ define texture modulation $T$. A multiplicative composition model combines $B$ and $T$ to reconstruct the input palmprint.}
	\label{fig:introduction}
\end{figure}

Palmprint recognition relies on the distinctive spatial patterns of palm lines and wrinkles to distinguish individuals~\cite{zhang2003online,yang2023multiorder}. In captured images, these textures appear on palm surfaces with varying color and illumination. Although existing recognition methods effectively extract discriminative texture features~\cite{gao2026palmprintsurvey}, they do not explicitly describe how these textures interact with the base appearance to form the observed image. We therefore seek an image-level representation that lets us examine and edit palmprint texture and base appearance separately and reconstruct the observed image through an explicit composition rule.

Palmprint synthesis and manipulation methods offer control over generated or transformed images~\cite{shao2019palmgan,shao2021jpfa,shen2023rpgpalm,jin2025diffpalm,jin2025uaa}, but do not directly decompose a real palmprint into editable factors. Content--style editing methods separate latent codes and rely on learned decoders to recombine them~\cite{huang2018munit,park2020swapping}. Intrinsic image decomposition explicitly models reflectance and illumination~\cite{bell2014intrinsic,careaga2023ordinal}, yet these components are not designed to isolate discriminative palmprint texture from base appearance.

To address this gap, we introduce \emph{Palmprint Decomposer}, an interpretable framework based on a fixed multiplicative composition model (Figure~\ref{fig:introduction}). It represents a palmprint image through \emph{base appearance} and \emph{texture modulation}, each characterized by two factors. A color map describes the underlying palm color, while a light map captures spatial brightness variations; together, they characterize the base appearance. A structure map captures the spatial patterns of palm lines and wrinkles, while a global texture strength parameter scales these patterns to determine texture modulation. The composition model reconstructs the image by locally attenuating the base appearance through texture modulation. This separation allows palm color and brightness to be adjusted independently of texture modulation, while the structure map and texture strength can be modified without affecting the base appearance.

Although the composition model explicitly defines how factors are combined, accurate reconstruction alone does not guarantee meaningful separation of palmprint texture from base appearance. Fine creases may still remain in the estimated base appearance even when the reconstructed image closely matches the input. To address this ambiguity in the absence of real-image factor annotations, we train the Factor Decomposer (FD) exclusively on synthetic data with direct factor supervision. Asymmetric decoding paths limit fine texture leakage into the base appearance while preserving full-resolution texture information in the structure map, and decomposition constraints further guide factor separation. To improve reconstruction fidelity on real palmprints, Instance-Wise Refinement (IWR) directly optimizes the predicted factors for each input without further involving the network.

We evaluate the proposed representation in terms of reconstruction fidelity, factor separation, and discriminative texture preservation. Experiments across seven datasets demonstrate high reconstruction fidelity, with the extracted structure maps generally achieving verification performance comparable to the original palmprints. Compared with existing methods, our approach achieves better factor separation, with less texture leakage into the base appearance while preserving discriminative texture information. Cross-dataset recomposition experiments further support the editability of the representation, with detailed results provided in the supplementary material.

Our main contributions are:
\begin{itemize}
	\item To the best of our knowledge, this is the first palmprint-specific image model to explicitly separate texture modulation from base appearance using four interpretable factors, enabling independent editing and recomposition through a simple and explicit composition model.

	\item We develop a decomposition framework trained exclusively on synthetic palmprints, employing asymmetric decoding and decomposition constraints to suppress texture leakage, together with IWR to improve reconstruction fidelity on real inputs.
	
	\item Extensive experiments across seven datasets demonstrate high reconstruction fidelity, effective factor separation, and preservation of discriminative texture information under diverse acquisition conditions, with additional experiments supporting the editability of the representation.
\end{itemize}

\section{Related Work}
\label{sec:related_work}

\subsection{Palmprint Recognition}

Palmprint recognition relies on discriminative patterns of palm lines and wrinkles. Classical methods encode Gabor phase and local orientations~\cite{zhang2003online,kong2004compcode,jia2008rloc,fei2016doc}, while learning-based approaches incorporate filtering and competition mechanisms into feature extraction~\cite{genovese2019palmnet,liang2021compnet,yang2023ccnet,yang2023co3net}. Multiple-order Texture Co-occurrence Code (MTCC)~\cite{yang2023multiorder} further characterizes first- and second-order texture responses. For cross-domain recognition, ISFDNet~\cite{liu2026isfdnet} separates identity and domain style in the feature space. These methods underscore the discriminative importance of palmprint textures, motivating their explicit representation through the structure map in our framework.

\subsection{Palmprint Synthesis and Manipulation}

Palmprint synthesis and manipulation have been explored to model variations in palm appearance, geometry, and texture. PalmGAN~\cite{shao2019palmgan} transfers palmprint appearance across domains, while JPFA~\cite{shao2021jpfa} integrates pixel-level transfer with feature alignment. Unified Adversarial Augmentation~\cite{jin2025uaa} combines identity-preserving generation with geometric transformations, and FlowPalm~\cite{zou2026flowpalm} models nonrigid variations through optical flow.

Other approaches incorporate crease patterns to guide synthesis. RPG-Palm~\cite{shen2023rpgpalm} uses B\'ezier creases with identity-aware supervision, while Diff-Palm~\cite{jin2025diffpalm} conditions diffusion on polynomial creases. PCE-Palm~\cite{jin2024pcepalm} introduces the Palm Crease Energy representation to characterize crease structures. Although these methods enable palmprint synthesis and manipulation, they do not explicitly decompose real palmprints into independently editable factors.

\subsection{Image Decomposition and Editing}

Intrinsic image decomposition separates images into reflectance and illumination~\cite{bell2014intrinsic}, with learning-based approaches leveraging paired exposures~\cite{wei2018retinex} or richer shading representations~\cite{careaga2023ordinal,careaga2024colorful}. Forsyth and Rock~\cite{forsyth2022paradigms} learn decomposition from products of sampled albedo and shading fields, supplemented by constraints on unlabeled real images. Huang et al.~\cite{huang2026paidnet} employ asymmetric reflectance and shading decoders with a shading regularizer to suppress unwanted texture. Structure--texture methods such as relative total variation~\cite{xu2012rtv} extract prominent image structures, while Hong et al.~\cite{hong2014vo} use Vese--Osher decomposition to obtain blur-stable structural features for palmprint recognition.

Editable representations further differ in how their components are recombined. MUNIT~\cite{huang2018munit} and Swapping Autoencoder~\cite{park2020swapping} rely on learned generators to recombine disentangled representations. RGB$\leftrightarrow$X~\cite{zeng2024rgbx} supports intrinsic-channel estimation and conditional synthesis, while IntrinsicEdit~\cite{lyu2025intrinsicedit} combines diffusion inversion with intrinsic-channel manipulation. In contrast, our framework explicitly separates palmprint texture modulation from base appearance, which captures color and spatial brightness variations. Trained exclusively on synthetic data, the Factor Decomposer estimates these factors, enabling direct editing and recomposition through a composition model.

\begin{figure}[!ht]
	\centering
	\includegraphics[width=0.98\linewidth]{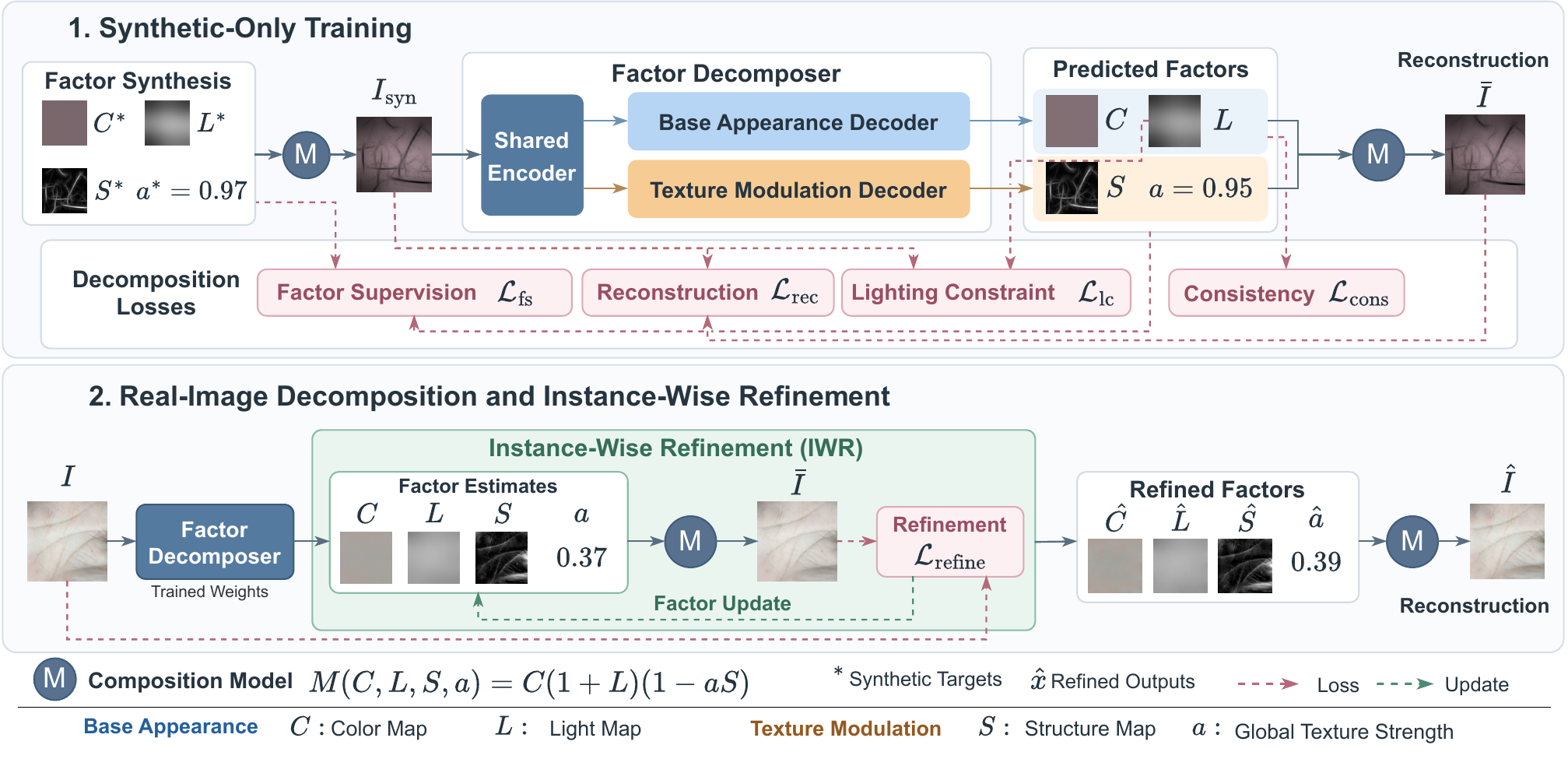}
	\caption{Overview of the proposed Palmprint Decomposer framework. During synthetic-only training, FD estimates four factors from synthetic images using asymmetric decoding and decomposition losses. For real palmprints, IWR directly optimizes the initial factor estimates to improve reconstruction fidelity.}
	\label{fig:overview}
\end{figure}

\section{Method}
\label{sec:method}

\subsection{Problem Formulation}
\label{sec:formulation}

Given a palmprint region of interest (ROI) $I\in[0,1]^{3\times H\times W}$, our goal is to represent it through interpretable, independently editable factors and an explicit composition model. We describe a palmprint image using two components: \emph{base appearance} $B$, which characterizes the underlying color and spatial brightness variations, and \emph{texture modulation} $T$, which controls the local attenuation associated with palmprint patterns.

The base appearance is defined by a \emph{color map} $C\in[0,1]^{3\times H\times W}$ and a \emph{light map} $L\in[-1,1]^{1\times H\times W}$. Texture modulation is determined by a \emph{structure map} $S\in[0,1]^{1\times H\times W}$, encoding the spatial patterns of palm lines and wrinkles, and a scalar \emph{global texture strength} $a\in[0,1]$, controlling their overall strength. The fixed composition model $M$ reconstructs the image from these factors:
\begin{equation}
\bar I=M(C,L,S,a)=C(1+L)(1-aS).
\label{eq:composition}
\end{equation}
Here, $B=C(1+L)$ and $T=aS$, giving the equivalent form $\bar I=B(1-T)$. All products are element-wise, with single-channel maps broadcast across RGB channels. The explicit composition model allows individual factors to be adjusted or exchanged for controlled palmprint recomposition.

The composition model provides a simple and explicit approximation of palmprint appearance, guided by visual and recognition-related considerations. The base appearance aims to minimize visible and recognizable palmprint texture, while texture modulation preserves discriminative information for recognition. The distinct factor roles and fixed composition rule enable interpretable decomposition and independent, predictable editing.

\subsection{Framework Overview}
\label{sec:overview}

Figure~\ref{fig:overview} illustrates the proposed framework, which consists of two stages: synthetic-only training and real-image decomposition with instance-wise refinement. The first stage learns factor decomposition from synthetic data, while the second adapts the estimated factors to individual real images.

During synthetic-only training, synthetic factors $(C^*,L^*,S^*,a^*)$ are combined through the composition model to generate synthetic images $I_{\mathrm{syn}}$. The Factor Decomposer (FD) employs a shared encoder and asymmetric decoders to estimate base appearance and texture modulation factors. The reconstruction loss promotes faithful image reconstruction, while factor supervision, consistency, and lighting constraints guide factor allocation and reduce texture leakage. For real palmprints, the trained FD predicts initial factors $(C,L,S,a)$. Starting from these estimates, Instance-Wise Refinement (IWR) directly optimizes the factors through the fixed composition model, without further involving the network. The refined factors $(\hat C,\hat L,\hat S,\hat a)$ yield the final reconstruction $\hat I=M(\hat C,\hat L,\hat S,\hat a)$.

\subsection{Factor Synthesis}
\label{sec:synthesis}

\begin{figure}[t]
	\centering
	\includegraphics[width=0.63\linewidth]{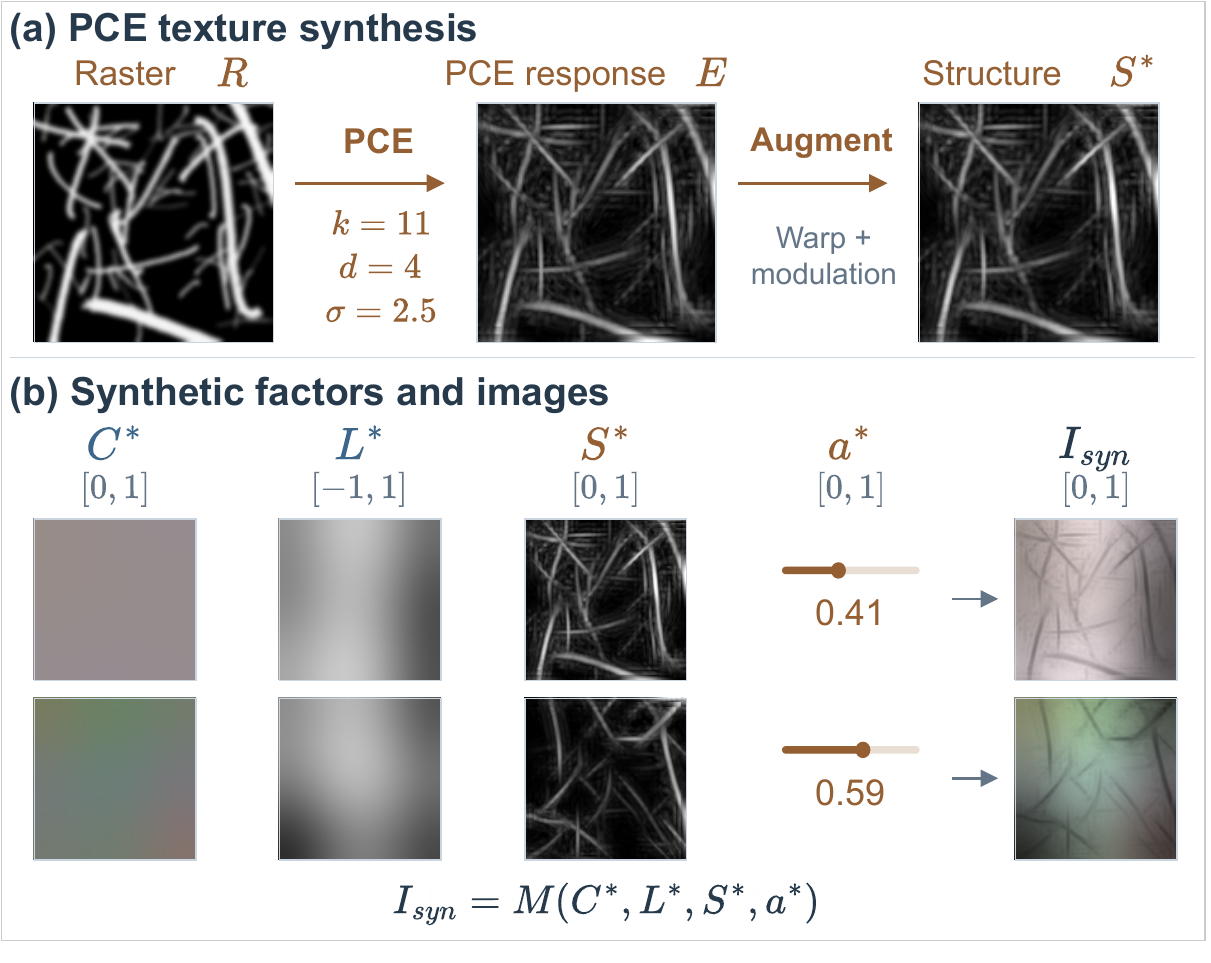}
	\caption{Synthetic factor generation and composition. (a) B\'ezier curves are rasterized and converted into PCE responses, then augmented to generate structure maps. (b) Examples of synthetic factors and their composed images. }
	\label{fig:factor_synthesis}
\end{figure}

Figure~\ref{fig:factor_synthesis} illustrates the texture synthesis process and provides examples of the resulting synthetic factors. The synthesis procedure is described below.

\paragraph{Base Appearance Synthesis.}
Base appearance synthesis assigns spatial color variations to the color map $C^*$ and brightness variations to the light map $L^*$. We perturb a sampled base color using low-frequency discrete cosine transform (DCT) bases while preserving its luminance. A separate normalized DCT field $F$ defines $L^*=\mu+\alpha F$, where the offset $\mu$ and amplitude $\alpha$ control overall brightness and spatial contrast, respectively. Only samples whose composed base appearance lies within $[0,1]$ is adopted.

\paragraph{Texture Modulation Synthesis.}
To simulate realistic palmprint patterns, we generate multiscale texture structures using coarse, medium, and fine B\'ezier curves with varying lengths, widths, orientations, and curvatures. Random branches and crossings enhance geometric diversity, while randomized blurring smooths the rasterized lines. Furthermore, we adopt the Palm Crease Energy (PCE) representation from PCE-Palm~\cite{jin2024pcepalm} to enhance the synthesized textures. Specifically, local mean subtraction and Gaussian-weighted modified finite Radon transform (MFRAT) filtering~\cite{jia2008rloc,jin2024pcepalm} produce directional responses. The maximum response across orientations is retained at each pixel, followed by zero-clipping and normalization.

Flips, rotations, and smooth nonrigid deformations further diversify the spatial patterns, while spatially varying amplitude adjustments introduce local texture variations. The resulting structure map $S^*\in[0,1]$ captures these variations, whereas a separately sampled scalar $a^*\sim\mathcal U(0,1)$ controls the global texture strength. Finally, the synthetic image $I_{\mathrm{syn}}=M(C^*,L^*,S^*,a^*)$ serves as the training input, with the synthetic factors providing direct supervision for FD.

\subsection{Factor Decomposer}
\label{sec:decomposer}

To suppress texture leakage into the base appearance, FD adopts asymmetric decoding paths for base appearance and texture modulation. A shared encoder extracts multiscale features, which are processed by two dedicated decoders to estimate the corresponding factors. The encoder and both decoders employ residual blocks~\cite{he2016resnet}.

The base appearance decoder operates at half resolution and omits the skip connection from the encoder's highest-resolution stage to limit fine texture details. For a $128\times128$ input, it predicts $C$ and $L$ at $64\times64$, followed by bilinear upsampling. In contrast, the texture modulation decoder retains skip connections at all encoder scales, following U-Net~\cite{ronneberger2015unet}, and produces full-resolution features to capture fine palmprint structures.

Dedicated prediction heads map the decoded features to the four factors. Separate $1\times1$ convolutions predict $C$, $L$, and $S$ from their corresponding decoder features. Global average pooling followed by a linear layer predicts the global texture strength $a$ from the texture modulation features. The complete architecture, including encoder skip connections and factor prediction heads, is illustrated in Supplementary Section~\ref{sec:sup_synthesis}.

\subsection{Decomposition Losses}
\label{sec:losses}

Since reconstruction alone cannot determine how information should be allocated among the factors, additional supervision and constraints are needed to guide the decomposition. The decomposition losses combine factor supervision with reconstruction, consistency, and lighting constraints.

\paragraph{Factor Supervision.}
The synthetic factors provide direct supervision for the predicted color, light, and structure maps, as well as the global texture strength:
\begin{equation}
	\mathcal L_{\mathrm{fs}}=
	\sum_{F\in\{C,L,S\}}\|F-F^*\|_1+h_{\beta}(a-a^*),
	\label{eq:factor_supervision}
\end{equation}
where $\|\cdot\|_1$ denotes the mean absolute error and $h_{\beta}$ is the smooth $L_1$ loss with $\beta=0.05$.

\paragraph{Reconstruction.}
The reconstruction loss encourages the predicted factors to reproduce the synthetic input through the composition model:
\begin{equation}
	\mathcal L_{\mathrm{rec}}=
	\|M(C,L,S,a)-I_{\mathrm{syn}}\|_1.
	\label{eq:reconstruction_loss}
\end{equation}

\paragraph{Consistency.}
To suppress texture leakage into the base appearance, we construct synthetic image pairs with identical base appearance factors $(C^*,L^*)$ but different texture modulation factors $(S^*,a^*)$, and penalize discrepancies between their predicted color and light maps:
\begin{equation}
	\mathcal L_{\mathrm{cons}}=
	\tfrac12\left(
	\|C_i-C_j\|_1+\|L_i-L_j\|_1
	\right).
	\label{eq:consistency_loss}
\end{equation}

\paragraph{Lighting Constraint.}
The color map may absorb spatial brightness variations, leaving the light map nearly uniform. To discourage this, we compare the smoothed and mean-normalized representations of input luminance and the brightness multiplier $1+L$ derived from the predicted light map, denoted by $U_I$ and $U_L$, respectively. Smoothing is performed by averaging each field to $8\times8$ and upsampling it to the input resolution. We penalize insufficient spatial variation in $U_L$:
\begin{equation}
	\mathcal L_{\mathrm{lc}}=
	\bigl[\rho\,\sigma(U_I)-\sigma(U_L)\bigr]_+,
	\label{eq:lighting_constraint}
\end{equation}
where $\sigma$ denotes the spatial standard deviation, $[x]_+=\max(x,0)$, and $\rho=0.5$ sets a moderate lower bound relative to the input luminance variation. By discouraging degenerate lighting estimates, this constraint further guides factor allocation and contributes to improved factor separation. Supplementary Section~\ref{sec:sup_synthesis} provides further implementation details.

The total training objective combines the four terms:
\begin{equation}
	\mathcal L=
	\mathcal L_{\mathrm{fs}}+
	\mathcal L_{\mathrm{rec}}+
	\mathcal L_{\mathrm{cons}}+
	\mathcal L_{\mathrm{lc}},
	\label{eq:training_objective}
\end{equation}
with each term averaged over the corresponding samples or sample pairs.

\subsection{Instance-Wise Refinement}
\label{sec:refinement}

Since FD is trained exclusively on synthetic data, its initial factor estimates may not accurately reconstruct real palmprint images. IWR uses these estimates as initialization and directly optimizes the factors through the fixed composition model.

To suppress fine texture leakage into the base appearance during refinement, updates to $C$ and $L$ are parameterized as $16\times16$ residual fields, which are bicubically upsampled and added to their initial predictions. In contrast, the structure map $S$ is optimized at full resolution, while the global texture strength $a$ is optimized as a scalar. All factors are constrained to their respective ranges.

To further suppress texture leakage, we combine a reconstruction term with a response-overlap constraint:
\begin{equation}
	\mathcal L_{\mathrm{refine}}
	=100\,\operatorname{MSE}(\bar I,I)
	+0.1\,\mathcal L_{\mathrm{overlap}}(B,I),
	\label{eq:refinement_objective}
\end{equation}
where $\bar I=M(C,L,S,a)$ and $B=C(1+L)$ are computed from the current factor estimates. The overlap constraint penalizes shared directional texture responses between the base appearance and input image. Specifically, it averages the products of absolute first- and second-order Gabor responses, normalized by the squared root-mean-square magnitude of the corresponding input responses, with equal weight for both orders. We use 100 Adam~\cite{kingma2015adam} steps per image (Supplementary Section~\ref{sec:sup_refinement}). The resulting refined factors $(\hat C,\hat L,\hat S,\hat a)$ produce the final reconstruction $\hat I=M(\hat C,\hat L,\hat S,\hat a)$.

\begin{table}[t]
\centering
\definecolor{tablegain}{HTML}{1B7F3A}
\definecolor{tableloss}{HTML}{B23A3A}

\caption{FD + IWR results across all datasets and subsets. Sim. denotes reconstruction-to-input texture similarity; Intra/Inter denotes mean same-/different-identity similarities of original images; C/L/B--O gives the texture similarity of each of $C$, $L$, and $B$ to the original input. Original reports EER (\%) and $d'$, while Structure Map $S$ and Reconstruction report changes relative to Original. All similarities are computed using MTCC. \textcolor{tablegain}{Green}/\textcolor{tableloss}{red} denotes improvement/degradation; changes rounding to zero are black.}
\label{tab:multidataset_final}
\setlength{\tabcolsep}{3pt}
\renewcommand{\arraystretch}{1.12}
\scriptsize
\resizebox{\textwidth}{!}{
\begin{tabular}{lrrrrccrrrrrr}
\toprule
& \multicolumn{4}{c}{Reconstruction} & Original similarity & Separation & \multicolumn{2}{c}{Original} & \multicolumn{2}{c}{Structure Map $S$} & \multicolumn{2}{c}{Reconstruction} \\
\cmidrule(lr){2-5}\cmidrule(lr){6-6}\cmidrule(lr){7-7}\cmidrule(lr){8-9}\cmidrule(lr){10-11}\cmidrule(lr){12-13}
Dataset/subset & PSNR$\uparrow$ & SSIM$\uparrow$ & LPIPS$\downarrow$ & Sim.$\uparrow$ & Intra$\uparrow$/Inter$\downarrow$ & C/L/B--O$\downarrow$ & EER$\downarrow$ & $d'\uparrow$ & $\Delta$EER$\downarrow$ & $\Delta d'\uparrow$ & $\Delta$EER$\downarrow$ & $\Delta d'\uparrow$ \\
\midrule
IITD & 42.62 & 0.980 & 0.030 & 0.99 & 0.79 / 0.55 & 0.54 / 0.56 / 0.56 & 1.21 & 4.60 & \textcolor{tablegain}{$-0.05$} & \textcolor{tableloss}{$-0.03$} & \textcolor{tableloss}{$+0.02$} & \textcolor{tableloss}{$-0.01$} \\
PolyU & 48.38 & 0.999 & 0.003 & 0.99 & 0.78 / 0.54 & 0.56 / 0.55 / 0.57 & 0.10 & 5.67 & \textcolor{tableloss}{$+0.01$} & \textcolor{tablegain}{$+0.01$} & $0.00$ & \textcolor{tablegain}{$+0.02$} \\
PolyU\_CF & 43.19 & 0.974 & 0.020 & 0.97 & 0.77 / 0.56 & 0.56 / 0.53 / 0.58 & 2.40 & 3.94 & \textcolor{tableloss}{$+0.03$} & \textcolor{tableloss}{$-0.09$} & \textcolor{tableloss}{$+0.07$} & \textcolor{tableloss}{$-0.01$} \\
Tongji & 42.84 & 0.981 & 0.032 & 0.98 & 0.77 / 0.55 & 0.55 / 0.55 / 0.58 & 1.11 & 4.07 & \textcolor{tablegain}{$-0.22$} & \textcolor{tablegain}{$+0.06$} & \textcolor{tablegain}{$-0.01$} & \textcolor{tablegain}{$+0.01$} \\
Zhou & 46.40 & 0.989 & 0.022 & 0.98 & 0.80 / 0.54 & 0.55 / 0.55 / 0.57 & 2.49 & 3.95 & \textcolor{tablegain}{$-0.33$} & \textcolor{tablegain}{$+0.02$} & $0.00$ & \textcolor{tablegain}{$+0.02$} \\
XJTU-UP\_HF & 44.97 & 0.985 & 0.017 & 0.96 & 0.78 / 0.54 & 0.50 / 0.53 / 0.65 & 0.81 & 4.61 & \textcolor{tableloss}{$+0.68$} & \textcolor{tableloss}{$-0.57$} & \textcolor{tableloss}{$+0.06$} & \textcolor{tableloss}{$-0.03$} \\
XJTU-UP\_HN & 52.36 & 0.996 & 0.007 & 0.98 & 0.77 / 0.54 & 0.52 / 0.54 / 0.54 & 1.33 & 4.33 & $0.00$ & \textcolor{tablegain}{$+0.01$} & \textcolor{tableloss}{$+0.02$} & $0.00$ \\
XJTU-UP\_IF & 45.48 & 0.988 & 0.022 & 0.98 & 0.79 / 0.54 & 0.52 / 0.53 / 0.62 & 1.15 & 4.52 & $0.00$ & \textcolor{tableloss}{$-0.05$} & \textcolor{tablegain}{$-0.01$} & $0.00$ \\
XJTU-UP\_IN & 49.06 & 0.994 & 0.013 & 0.98 & 0.77 / 0.54 & 0.52 / 0.54 / 0.55 & 1.48 & 4.12 & \textcolor{tablegain}{$-0.02$} & \textcolor{tablegain}{$+0.03$} & \textcolor{tablegain}{$-0.02$} & $0.00$ \\
CASIA\_460 & 59.19 & 0.999 & 0.003 & 0.99 & 0.71 / 0.54 & 0.52 / 0.56 / 0.55 & 2.20 & 3.65 & \textcolor{tablegain}{$-0.03$} & \textcolor{tablegain}{$+0.01$} & \textcolor{tablegain}{$-0.07$} & $0.00$ \\
CASIA\_630 & 59.39 & 0.999 & 0.004 & 0.99 & 0.74 / 0.56 & 0.53 / 0.56 / 0.56 & 2.10 & 3.88 & \textcolor{tablegain}{$-0.13$} & \textcolor{tablegain}{$+0.11$} & $0.00$ & $0.00$ \\
CASIA\_700 & 58.81 & 0.999 & 0.005 & 0.98 & 0.73 / 0.56 & 0.53 / 0.55 / 0.56 & 3.10 & 3.62 & \textcolor{tablegain}{$-0.77$} & \textcolor{tablegain}{$+0.15$} & \textcolor{tablegain}{$-0.07$} & $0.00$ \\
CASIA\_850 & 58.34 & 0.999 & 0.005 & 0.98 & 0.74 / 0.55 & 0.54 / 0.54 / 0.55 & 2.13 & 3.88 & \textcolor{tableloss}{$+0.07$} & $0.00$ & \textcolor{tablegain}{$-0.03$} & \textcolor{tablegain}{$+0.01$} \\
CASIA\_940 & 59.19 & 0.999 & 0.004 & 0.98 & 0.74 / 0.55 & 0.53 / 0.54 / 0.55 & 2.23 & 3.79 & \textcolor{tablegain}{$-0.17$} & \textcolor{tablegain}{$+0.07$} & \textcolor{tablegain}{$-0.03$} & $0.00$ \\
CASIA\_WHT & 58.43 & 0.999 & 0.004 & 0.99 & 0.72 / 0.55 & 0.52 / 0.56 / 0.56 & 1.99 & 3.71 & \textcolor{tablegain}{$-0.22$} & \textcolor{tablegain}{$+0.02$} & \textcolor{tablegain}{$-0.12$} & \textcolor{tablegain}{$+0.01$} \\
\bottomrule
\end{tabular}
}
\end{table}

\section{Experiments}
\label{sec:experiments}

\subsection{Experimental Setup}
\label{sec:experimental_setup}

\paragraph{Datasets.}
Experiments use seven palmprint datasets spanning controlled, mobile, and multispectral acquisition. IITD~\cite{kumar2008cohort} and Tongji~\cite{zhang2017contactless} provide contactless palmprint images under controlled conditions, whereas PolyU~\cite{zhang2003online} contains contact-based grayscale images. PolyU\_CF~\cite{kanhangad2011contactless} provides contact-free hand images. Zhou~\cite{zhou2019keypoint} and XJTU-UP~\cite{shao2019xjtu} contain images acquired with mobile devices under varying scene and illumination conditions; four XJTU-UP subsets (HF, HN, IF, IN) are used. CASIA~\cite{casiaMultispectral} provides six spectral subsets covering visible and near-infrared imaging. Together, these datasets yield 15 dataset or subset evaluation groups. Evaluation uses all available images in each selected dataset or subset.

\paragraph{Implementation Details.}
All ROIs are converted to RGB and resized to $128\times128$. FD is trained exclusively on synthetic images for 100 epochs, with 10,000 samples per epoch, a batch size of 120, and a learning rate of $2\times10^{-4}$.

All test sets use the same exponential moving average checkpoint. Unless otherwise stated, results are obtained with FD followed by 100 Adam steps of IWR per image, using an initial learning rate of 0.01 for all factors. Experiments run on an AMD Ryzen 9 7900X CPU and an NVIDIA RTX 4090 GPU. FD has 9.26\,M parameters and requires 12.21\,GFLOPs per forward pass. FD and 100-step IWR take 2.49\,ms and 203.15\,ms per image on average, respectively, for a total of 205.64\,ms.

\paragraph{Evaluation Protocol.}
Unless otherwise stated, factor symbols denote refined outputs, with hats omitted for brevity. MTCC~\cite{yang2023multiorder} is used to measure texture similarity throughout the evaluation. Reconstruction fidelity is evaluated using PSNR, SSIM~\cite{wang2004ssim}, LPIPS~\cite{zhang2018lpips}, and texture similarity between the reconstruction and input. We measure the texture similarity of the color map $C$, the light map $L$, and their combined base appearance $B$ to the input. Lower values indicate less input texture retained in the respective output.

Discriminative texture preservation is evaluated using the equal error rate (EER) and decidability index $d'$~\cite{daugman2004iris}, derived from genuine and impostor distributions of MTCC-based texture similarity scores over all image pairs within each dataset or subset. We compare the structure map $S$ with the original and reconstructed images, while baseline comparisons evaluate each method's texture output ($T=aS$ for ours). The mean genuine and impostor similarities of the original images are reported as Intra and Inter, respectively. Editability is illustrated through independent factor editing and further explored through cross-dataset factor exchange, with detailed results provided in Supplementary Section~\ref{sec:factor_editing_exchange}.

\subsection{Decomposition on Real Palmprints}
\label{sec:real_decomposition}

\begin{figure}[t]
    \centering
    \includegraphics[width=\textwidth]{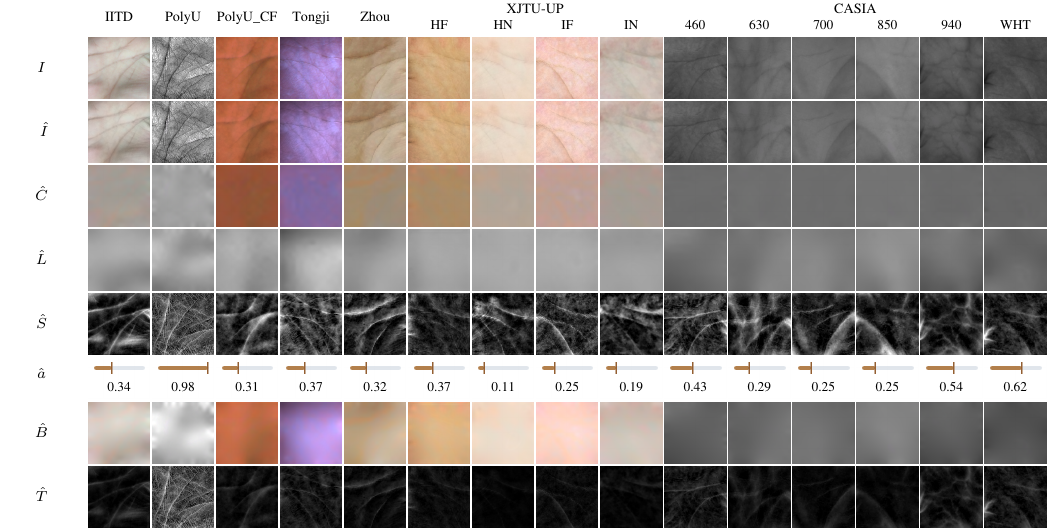}
    \caption{Decomposition on real palmprints. Original images, final reconstructions, refined factors, and the derived base appearance $\hat B=\hat C(1+\hat L)$ and texture modulation $\hat T=\hat a\hat S$ across datasets and subsets.}
    \label{fig:dataset_decomposition}
\end{figure}

Table~\ref{tab:multidataset_final} reports PSNR values of 42.62--59.39\,dB, SSIM of 0.974--0.999, and reconstruction-to-input texture similarities of 0.96--0.99. These high texture similarities indicate that the reconstructed images closely preserve the palmprint patterns of the original inputs. Overall, the results demonstrate that FD, trained exclusively on synthetic data, achieves high reconstruction fidelity on diverse real palmprints through IWR.

Beyond reconstruction fidelity, we assess texture retention in the estimated base appearance using MTCC similarity to the input, with original-image intra- and inter-identity similarities providing contextual references. The similarities of $C$, $L$, and $B$ to the input are consistently lower than the intra-identity similarities and generally close to the inter-identity similarities. In particular, B--O ranges from 0.54 to 0.58 in 13 of the 15 evaluation groups, compared with inter-identity similarities of 0.54--0.56. Together with the visualizations in Figure~\ref{fig:dataset_decomposition}, these results support reduced retention of input-specific palmprint texture in the estimated base appearance.

Both the reconstructed images and extracted structure maps generally achieve verification performance comparable to that of the original palmprints, indicating effective preservation of discriminative information. The reconstructions exhibit only minor changes in EER and $d'$, while the structure maps sometimes outperform the originals, with EER reductions of 0.22 and 0.33 percentage points on Tongji and Zhou, respectively, and improvements on five of the six CASIA subsets. On XJTU-UP\_HF, the structure-map EER increases by 0.68 percentage points and $d'$ decreases by 0.57. This may stem from the gap between our simplified palmprint formation model and the complex interactions between texture and appearance under certain acquisition conditions, making accurate reconstruction and factor separation more challenging. Overall, these results demonstrate that the proposed representation achieves high reconstruction fidelity and effective factor separation while preserving discriminative palmprint texture information.

\begin{figure}[t]
    \centering
    \includegraphics[width=0.85\linewidth]{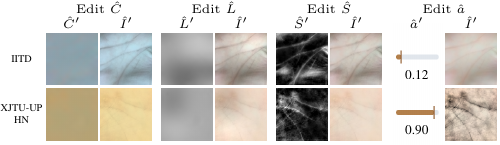}
    \caption{Independent factor editing. Each group shows an edited factor and the resulting recomposed image, with the other three factors fixed. Primes denote the edited factors and recomposed images.}
    \label{fig:independent_factor_editing}
\end{figure}

Figure~\ref{fig:independent_factor_editing} illustrates independent editing of color, spatial brightness, spatial texture patterns, and global texture strength while holding the other factors fixed. Cross-dataset factor exchange experiments further demonstrate the editability of the representation, showing that texture and base appearance can be independently manipulated while preserving their respective characteristics. Detailed results are provided in Supplementary Section~\ref{sec:factor_editing_exchange}.

\subsection{Comparison with Decomposition Baselines}
\label{sec:decomposition_baselines}

\begin{table}[tb]
\centering
\caption{Comparison of factor separation and texture verification on IITD and XJTU-UP. $B$ denotes the base appearance estimate (the filtered image for Gaussian and RTV), and $O$ denotes the original input. Hist. corr. measures mean RGB-channel histogram correlation. N/A indicates that only the texture output is evaluated. Bold marks the best result per metric and dataset.}
\label{tab:decomposition_baselines}
\setlength{\tabcolsep}{2pt}
\renewcommand{\arraystretch}{1.05}
\scriptsize
\begin{tabular*}{\columnwidth}{@{\extracolsep{\fill}}lrrrr@{}}
\toprule
& \multicolumn{2}{c}{Base appearance vs.\ input} & \multicolumn{2}{c}{Texture verification} \\
\cmidrule(lr){2-3}\cmidrule(l){4-5}
Method & Hist. corr.$\uparrow$ & B--O$\downarrow$ & EER (\%)$\downarrow$ & $d^{\prime}\uparrow$ \\
\midrule
\multicolumn{5}{l}{\textbf{IITD}} \\
Gaussian & \textbf{0.980} & 0.910 & 1.56 & 3.87 \\
RTV~\cite{xu2012rtv} & 0.826 & 0.788 & 1.33 & 4.35 \\
Direct PCE~\cite{jin2024pcepalm} & N/A & N/A & 6.32 & 3.02 \\
VO~\cite{hong2014vo} & N/A & N/A & 2.59 & 3.26 \\
FD & 0.676 & \textbf{0.521} & 1.20 & \textbf{4.61} \\
FD + IWR & 0.657 & 0.564 & \textbf{1.16} & 4.58 \\
\midrule
\multicolumn{5}{l}{\textbf{XJTU-UP}} \\
Gaussian & \textbf{0.966} & 0.903 & 3.46 & 2.98 \\
RTV & 0.818 & 0.764 & 3.22 & 3.15 \\
Direct PCE & N/A & N/A & 17.43 & 1.78 \\
VO & N/A & N/A & 5.02 & 2.45 \\
FD & 0.531 & \textbf{0.516} & \textbf{3.19} & \textbf{3.21} \\
FD + IWR & 0.555 & 0.590 & 3.50 & 3.07 \\
\bottomrule
\end{tabular*}
\end{table}

\begin{figure}[t]
    \centering
    \includegraphics[width=0.70\linewidth]{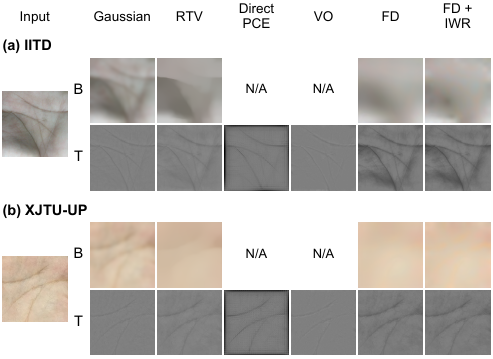}
    \caption{Visual comparison of base appearance estimates and texture outputs on IITD and XJTU-UP. For our method, $B$ and $T$ denote base appearance and texture modulation, respectively. Texture outputs are visualized with zero response mapped to mid-gray for consistent comparison. For Direct PCE and VO, only the texture outputs are shown.}
    \label{fig:decomposition_comparison}
\end{figure}

On IITD and XJTU-UP, we examine how well each method's texture output preserves discriminative palmprint patterns and, where a base appearance estimate is available, how much input texture it retains (Table~\ref{tab:decomposition_baselines} and Figure~\ref{fig:decomposition_comparison}). For Gaussian and relative total variation (RTV) filtering, we use the filtered image as the base appearance estimate $B$ and the luminance residual $Y(I-B)$ for texture matching. For Direct PCE~\cite{jin2024pcepalm} and grayscale Vese--Osher (VO) decomposition~\cite{hong2014vo}, we evaluate the crease response and the oscillatory component $v$, respectively. For our method, we evaluate base appearance $B$ and texture modulation $T=aS$ before and after IWR. Implementation details are given in Supplementary Section~\ref{sec:sup_baselines}.

At the settings reported in Table~\ref{tab:decomposition_baselines}, Gaussian and RTV yield higher RGB histogram correlations with the input but also higher B--O scores than FD. On XJTU-UP, FD achieves a B--O score of 0.516 compared with 0.764 for RTV, with similar texture EERs (3.19\% and 3.22\%, respectively). IWR improves reconstruction fidelity. After refinement, B--O increases on both datasets, and texture EER decreases on IITD but increases on XJTU-UP. Although these methods can extract base appearance or palmprint texture to varying degrees, our framework provides a finer-grained factorization with improved factor separation and preservation of discriminative texture information. Furthermore, the composition model explicitly defines the roles and interactions of these factors, enabling interpretable reconstruction and controlled editing.

\subsection{Ablation Studies}
\label{sec:ablations}

\begin{table}[!htbp]
\centering
\caption{Ablations on the complete IITD and PolyU sets. Each training variant independently modifies FD. FD + IWR reuses the 100-step final results from Table~\ref{tab:multidataset_final}, which also defines the reconstruction metrics and C/L/B--O. Structure Map $S$ and Recon. cells report EER (\%)/$d^{\prime}$. Bold marks the best displayed values within each dataset, including ties.}
\label{tab:ablation_final}
\setlength{\tabcolsep}{0.7pt}
\renewcommand{\arraystretch}{1.12}
\scriptsize
\begin{tabular*}{\columnwidth}{@{\extracolsep{\fill}}lrrrrccc@{}}
\toprule
& \multicolumn{4}{c}{Reconstruction} & Separation & \shortstack{Structure\\Map $S$} & Recon. \\
\cmidrule(lr){2-5}\cmidrule(lr){6-6}\cmidrule(lr){7-7}\cmidrule(lr){8-8}
Variant & PSNR$\uparrow$ & SSIM$\uparrow$ & LPIPS$\downarrow$ & Sim.$\uparrow$ & C/L/B--O$\downarrow$ & EER$\downarrow$/$d^{\prime}\uparrow$ & EER$\downarrow$/$d^{\prime}\uparrow$ \\
\midrule
\multicolumn{8}{l}{\textbf{IITD}} \\
\midrule
FD & 38.45 & 0.971 & 0.081 & 0.96 & 0.49/\textbf{0.53}/\textbf{0.52} & 1.20/\textbf{4.61} & 1.26/4.59 \\
w/o PCE & 36.61 & 0.961 & 0.104 & 0.94 & 0.50/0.57/0.57 & 1.23/4.53 & 1.26/4.53 \\
Full-res. Base & 39.52 & 0.976 & 0.065 & 0.97 & 0.53/0.58/0.59 & 1.23/4.60 & 1.25/4.57 \\
w/o Cons. & 38.20 & 0.973 & 0.074 & 0.96 & \textbf{0.45}/0.60/0.58 & 1.23/4.59 & 1.26/4.55 \\
w/o LC & 37.51 & 0.971 & 0.086 & 0.96 & 0.57/0.56/0.57 & 1.20/4.60 & \textbf{1.23}/4.58 \\
FD + IWR & \textbf{42.62} & \textbf{0.980} & \textbf{0.030} & \textbf{0.99} & 0.54/0.56/0.56 & \textbf{1.16}/4.58 & \textbf{1.23}/\textbf{4.60} \\
\addlinespace
\multicolumn{8}{l}{\textbf{PolyU}} \\
\midrule
FD & 31.13 & 0.973 & 0.036 & 0.96 & \textbf{0.51}/\textbf{0.51}/\textbf{0.51} & \textbf{0.10}/\textbf{5.82} & 0.11/\textbf{5.77} \\
w/o PCE & 28.23 & 0.916 & 0.102 & 0.94 & 0.54/0.56/0.56 & \textbf{0.10}/5.70 & \textbf{0.10}/5.66 \\
Full-res. Base & 29.88 & 0.971 & 0.037 & 0.97 & 0.54/0.55/0.56 & \textbf{0.10}/5.80 & \textbf{0.10}/5.75 \\
w/o Cons. & 31.75 & 0.977 & 0.037 & 0.96 & 0.62/0.65/0.66 & \textbf{0.10}/5.79 & 0.11/5.73 \\
w/o LC & 30.38 & 0.970 & 0.039 & 0.97 & 0.60/0.54/0.56 & \textbf{0.10}/5.78 & \textbf{0.10}/5.73 \\
FD + IWR & \textbf{48.38} & \textbf{0.999} & \textbf{0.003} & \textbf{0.99} & 0.56/0.55/0.57 & 0.11/5.68 & \textbf{0.10}/5.69 \\
\addlinespace
\bottomrule
\end{tabular*}
\par\smallskip
\begin{minipage}{\columnwidth}
\scriptsize
Cons.: Consistency; LC: Lighting Constraint. w/o PCE uses rasterized curves. Full-res. Base restores full-resolution base appearance decoding and the finest-scale skip connection.
\end{minipage}
\end{table}

Table~\ref{tab:ablation_final} compares independent variants of FD and the effects of IWR on IITD and PolyU. Replacing PCE responses with rasterized curves (w/o PCE) reduces reconstruction fidelity on real inputs. On PolyU, SSIM drops from 0.973 to 0.916, while LPIPS increases from 0.036 to 0.102. B--O also increases and structure-map $d'$ decreases on both datasets, suggesting that PCE responses help bridge the synthetic--real texture domain gap and improve the transfer of factor supervision.

Decoder and loss ablations further examine factor separation. Restoring full-resolution base appearance decoding and the finest-scale skip connection (Full-res. Base) increases B--O on both datasets. On IITD, B--O rises from 0.52 to 0.59 despite improved SSIM, supporting the use of restricted base appearance decoding to suppress texture leakage. Removing consistency (w/o Cons.) raises B--O from 0.51 to 0.66 on PolyU, demonstrating its role in reducing the sensitivity of base appearance to texture variations. Removing the lighting constraint (w/o LC) increases C--O from 0.49 to 0.57 on IITD and from 0.51 to 0.60 on PolyU, indicating that guiding color--light allocation also helps suppress texture leakage into the color map.

IWR substantially improves all reconstruction metrics on both datasets, while introducing only modest increases in B--O and minor variations in structure-map verification performance. These results demonstrate that IWR achieves a favorable balance between reconstruction fidelity, factor separation, and discriminative texture preservation. Supplementary Section~\ref{sec:sup_iwr} further examines the effects of refinement steps and the overlap constraint.

\FloatBarrier
\section{Conclusion}
\label{sec:conclusion}

We have presented Palmprint Decomposer, an interpretable and editable palmprint factorization framework based on an explicit composition model. Synthetic-only training with factor supervision and decomposition constraints enables meaningful factor separation, while Instance-Wise Refinement directly optimizes the predicted factors for improved real-image reconstruction without further involving the network. Experiments across seven datasets demonstrate high reconstruction fidelity, effective factor separation, and preservation of discriminative texture information. Future work will explore more expressive factor representations and composition models that better approximate real palmprint image formation.

\FloatBarrier
{\small
\bibliographystyle{unsrtnat}
\bibliography{main}
}
\clearpage
\appendix
\section*{Supplementary Material}
\setcounter{figure}{0}
\setcounter{table}{0}
\renewcommand{\thefigure}{S\arabic{figure}}
\renewcommand{\thetable}{S\arabic{table}}
\renewcommand{\theHfigure}{supp.\arabic{figure}}
\renewcommand{\theHtable}{supp.\arabic{table}}
\section{Synthesis, Architecture, and Training}
\label{sec:sup_synthesis}
This section provides implementation details for factor synthesis, the Factor Decomposer (FD), and synthetic-only training. Synthetic factors provide direct supervision, while paired images sharing the same base appearance factors support the consistency loss.

\paragraph{Base Appearance Synthesis.}
We sample the base luminance uniformly from $[0.4,0.6]$ and use a grayscale base with probability 0.2. For color samples, we add two spatial chromatic perturbations with amplitudes drawn uniformly from $[0,0.08]$. For both color and lighting, each discrete cosine transform (DCT) field uses the five modes with $p,q\geq 0$ and $1\leq p+q\leq 2$. We draw independent standard normal coefficients, subtract the spatial mean, and divide by the maximum absolute value with a floor of $10^{-6}$. Projecting each color perturbation onto the plane orthogonal to the RGB luminance weights varies chromaticity while preserving the base luminance. A separate normalized DCT field $F$ controls relative lighting through $L^*=\mu+\alpha F$, with offset $\mu\in[-0.2,0.2]$ and amplitude $\alpha\in[0.3,0.7]$. We retain samples whose base appearance $C^*(1+L^*)$ lies in $[0,1]$.

\paragraph{Texture Modulation Synthesis.}
The coarse, medium, and fine families contain 2--4, 8--18, and 24--48 primary cubic B\'ezier curves, respectively. Their nominal widths are 4--7, 2--3, and 1 pixel at output resolution. Counts and widths are sampled uniformly over integers. Before cropping, the endpoint separations span 60--90\%, 24--52\%, and 8--24\% of the image width, respectively. We rasterize at $3\times$ resolution, apply a Gaussian blur with radius drawn uniformly from 0.3--1.8 output pixels, and downsample bilinearly. Palm Crease Energy (PCE)~\cite{jin2024pcepalm} converts the rasterized curves into a continuous response map. We subtract a $9\times9$ local mean and apply Gaussian-weighted $11\times11$ directional kernels with dilation 4 and standard deviation 2.5 in kernel coordinates. The 12 orientations are uniformly spaced over $[0,\pi)$. At each pixel, we retain the maximum response over orientations, set negative responses to zero, and normalize the map by its spatial maximum with a numerical floor. Flips, rotations, smooth nonrigid deformations, and spatially varying amplitude adjustments produce the structure map $S^*$. Sampling $a^*$ uniformly from $[0,1]$ then varies global texture strength independently of the spatial pattern.

\begin{figure}[htbp]
    \centering
    \includegraphics[width=0.95\linewidth]{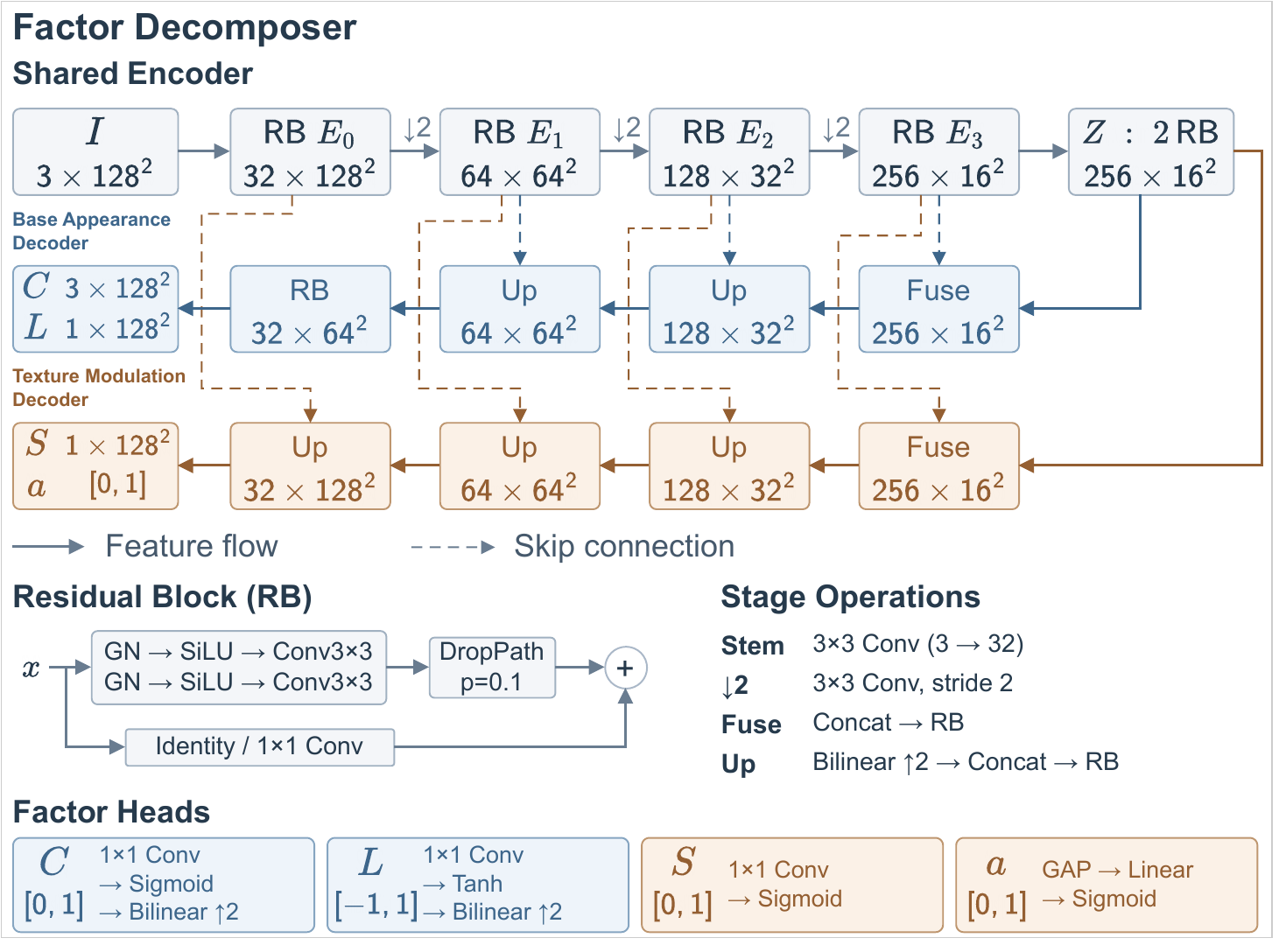}
    \caption{Factor Decomposer. The base appearance decoder operates at half resolution and omits the skip connection from the highest-resolution encoder stage. Its two heads predict the color map $C$ and light map $L$, followed by bilinear upsampling. The texture modulation decoder retains skip connections at all encoder scales. Its spatial head predicts the structure map $S$, while global average pooling followed by a linear layer predicts the global texture strength $a$. RB denotes residual block, GN group normalization, and GAP global average pooling.}
    \label{fig:sup_network}
\end{figure}

\paragraph{FD Architecture.}
Figure~\ref{fig:sup_network} specifies the encoder stages, decoder operations, and prediction heads of FD. The asymmetric decoding paths suppress fine texture leakage into base appearance while retaining full-resolution features for texture modulation. The diagram also specifies feature dimensions and residual-block operations.

\paragraph{Consistency and Lighting Constraints.}
To form a training pair, we reuse $C^*$ and $L^*$ and change $S^*$ and $a^*$. The consistency loss penalizes differences between the predicted color and light maps, encouraging base appearance estimates to remain insensitive to texture variation. The lighting constraint addresses a different ambiguity: a spatially varying color map could absorb lighting changes while $L$ remains nearly uniform. To discourage this allocation, we compare input luminance with the brightness multiplier $1+L$ derived from the predicted light map at the same coarse scale. Let $\mathcal P$ average its input to $8\times8$ and bilinearly upsample it to the ROI resolution. Let $\mathcal R(X)=X/\max(\langle X\rangle,10^{-6})-1$, where $\langle X\rangle$ is the spatial mean. The two fields, with $Y$ denoting luminance, are defined as:
\[
 U_I=\mathcal R(\mathcal P(Y(I_{\rm syn}))),\quad
 U_L=\mathcal P(\mathcal R(1+L)).
\]
We compute their spatial standard deviations with $10^{-8}$ added to the variance. A penalty applies when $\sigma(U_L)<0.5\,\sigma(U_I)$, without requiring the two fields to match point by point.

\paragraph{Training.}
FD is trained exclusively on synthetic images for 100 epochs, with 10,000 samples per epoch and a batch size of 120. We use AdamW with a learning rate of $2\times10^{-4}$, weight decay of $10^{-4}$, and gradient norm clipping at 1. The training seed is 1234. An exponential moving average of the network weights is updated after each optimizer step with decay 0.99.

\FloatBarrier
\section{Instance-Wise Refinement}
\label{sec:sup_refinement}
Starting from the factor estimates predicted by FD, Instance-Wise Refinement (IWR) directly optimizes the factors through the fixed composition model. Unless otherwise stated, IWR uses 100 Adam~\cite{kingma2015adam} steps per image.

\paragraph{Refinement Settings.}
We initialize the $16\times16$ residual fields for $C$ and $L$ at zero and upsample them bicubically with \texttt{align\_corners=False}. We clamp $C$ to $[10^{-4},1-10^{-4}]$, $L$ to $[-1,1]$, and $S,a$ to $[0,1]$. For all four factors, the Adam learning rate follows a half cosine from 0.01 at the first update to 0.001 at the last. We clip the joint gradient norm at 1.

\paragraph{Response-Overlap Constraint.}
For the response-overlap constraint, we filter luminance with six zero-mean $17\times17$ Gabor kernels at $30^\circ$ intervals starting at $0^\circ$ ($\omega=0.5$, $\kappa=2$). Applying the same kernel a second time gives the second-order response. Let $R_{k,\theta}(X)$ denote the response of order $k$ and orientation $\theta$ to image $X$, and let $\langle\cdot\rangle$ average over the ROI excluding a 16-pixel border. With $s_{k,\theta}=\max(\sqrt{\langle R_{k,\theta}(I)^2\rangle},10^{-6})$, the per-image penalty is defined as:
\[
\mathcal L_{\rm overlap}=\frac{1}{12}\sum_{k=1}^{2}\sum_{\theta}
\frac{\langle |R_{k,\theta}(B)R_{k,\theta}(I)|\rangle}{s_{k,\theta}^2}.
\]

\section{Evaluation Datasets}
\label{sec:sup_datasets}
Table~\ref{tab:sup_datasets} lists the real palmprints used for evaluation; none are used to train FD. Each identity denotes one palm. All inputs are $128\times128$ RGB ROIs, and the same exponential moving average checkpoint is used across acquisition conditions.

\begin{table}[ht]
    \centering
    \caption{Evaluated ROIs and palm identities. Each palm is counted once across acquisition conditions. Verification uses all unordered pairs of distinct images within each evaluation group.}
    \label{tab:sup_datasets}
    \small
    \begin{tabular*}{\columnwidth}{@{\extracolsep{\fill}}lrr@{}}
        \toprule
        Dataset & Images & Palm identities \\
        \midrule
        IITD & 2,600 & 460 \\
        Tongji & 12,000 & 600 \\
        PolyU & 7,752 & 386 \\
        PolyU\_CF & 1,770 & 177 \\
        Zhou & 1,295 & 166 \\
        XJTU-UP & 7,880 & 200 \\
        CASIA & 7,200 & 200 \\
        \bottomrule
    \end{tabular*}
\end{table}

\paragraph{Acquisition.}
IITD~\cite{kumar2008cohort} and Tongji~\cite{zhang2017contactless} use controlled contactless acquisition. Tongji uses a CCD camera with ring illumination and includes two acquisition sessions. PolyU~\cite{zhang2003online} provides contact-based grayscale images. PolyU\_CF~\cite{kanhangad2011contactless} pairs range scans with intensity images of the right hand. We use the 2D intensity images from both sessions. Zhou~\cite{zhou2019keypoint} represents mobile acquisition. The Zhou evaluation list used here contains 1,295 ROIs from 166 palms.

XJTU-UP~\cite{shao2019xjtu} varies both the phone and the illumination. HF and HN denote Huawei images under flash and natural lighting, respectively, and IF and IN denote the corresponding iPhone images. The evaluated conditions contain 1,980, 1,970, 1,960, and 1,970 images, respectively. CASIA~\cite{casiaMultispectral} contains 1,200 images per band at 460, 630, 700, 850, and 940\,nm and under white light.

\paragraph{Evaluation Groups.}
Table~\ref{tab:multidataset_final} evaluates 15 groups: five individual datasets, four XJTU-UP acquisition conditions, and six CASIA spectral subsets. Table~\ref{tab:decomposition_baselines} evaluates all 2,600 IITD images and all 7,880 XJTU-UP images at the dataset level. Verification uses all unordered pairs of distinct images within each group, including cross-condition pairs for dataset-level XJTU-UP evaluation.

\section{Decomposition Baselines and Metrics}
\label{sec:sup_baselines}
Table~\ref{tab:decomposition_baselines} evaluates three complementary properties: agreement with the input intensity distribution, input texture retained in the base appearance estimate, and discriminative information retained in the texture output. This section specifies the baseline outputs and evaluation metrics. For our method, the corresponding outputs are base appearance $B$ and texture modulation $T$.

\paragraph{Baseline Outputs.}
Gaussian filtering smooths each RGB channel with a standard deviation of 3 pixels, a radius of 9 pixels, and reflected boundaries. Relative total variation (RTV)~\cite{xu2012rtv} smooths the input using a direct solver with $\lambda=0.01$, initial spatial scale 3, sharpness 0.02, and four iterations. Both methods use the smoothed image as $B$. Subtracting $B$ from the input gives a signed residual whose luminance records details removed by smoothing.

Direct PCE~\cite{jin2024pcepalm} applies the directional filters described in Section~\ref{sec:sup_synthesis} to $1-Y(I)$ so that dark creases produce positive responses $P$. Our numerical implementation of Vese--Osher (VO) decomposition follows the published grayscale formulation~\cite{hong2014vo}. For VO, we evaluate the oscillatory component $v=\operatorname{div}g$ as the texture output. The original method uses the structure component for recognition. Only texture results are reported for VO and Direct PCE.

\paragraph{Structure Maps and Texture Modulation.}
Tables~\ref{tab:multidataset_final} and~\ref{tab:ablation_final} evaluate the structure map $S$, while the baseline comparison in Table~\ref{tab:decomposition_baselines} evaluates texture modulation $T=aS$. For baseline texture matching, signed outputs $r=Y(I-B)$ or $v$ are mapped to $(1+r)/2$, and nonnegative outputs $q=P$ or $T$ are mapped to $(1-q)/2$, followed by clipping to $[0,1]$.

\paragraph{Evaluation Metrics.}
For each base appearance estimate $B$ and its corresponding input $O$, we compute a 256-bin histogram per RGB channel for each image and average the three channel-wise correlations. This measures agreement in intensity distributions. To measure whether recognizable patterns persist at the same locations, we compare the aligned $32\times32$ MTCC templates~\cite{yang2023multiorder} of $B$ and $O$ with zero shift (B--O). For texture verification across different images, we allow shifts of up to two template cells and compute EER and the decidability index $d'$~\cite{daugman2004iris} from the genuine and impostor MTCC scores. A higher $d'$ indicates greater separation between the two score distributions.

\section{Factor Editing and Exchange}
\label{sec:factor_editing_exchange}

\begin{figure}[!t]
    \centering
    \includegraphics[width=0.68\linewidth]{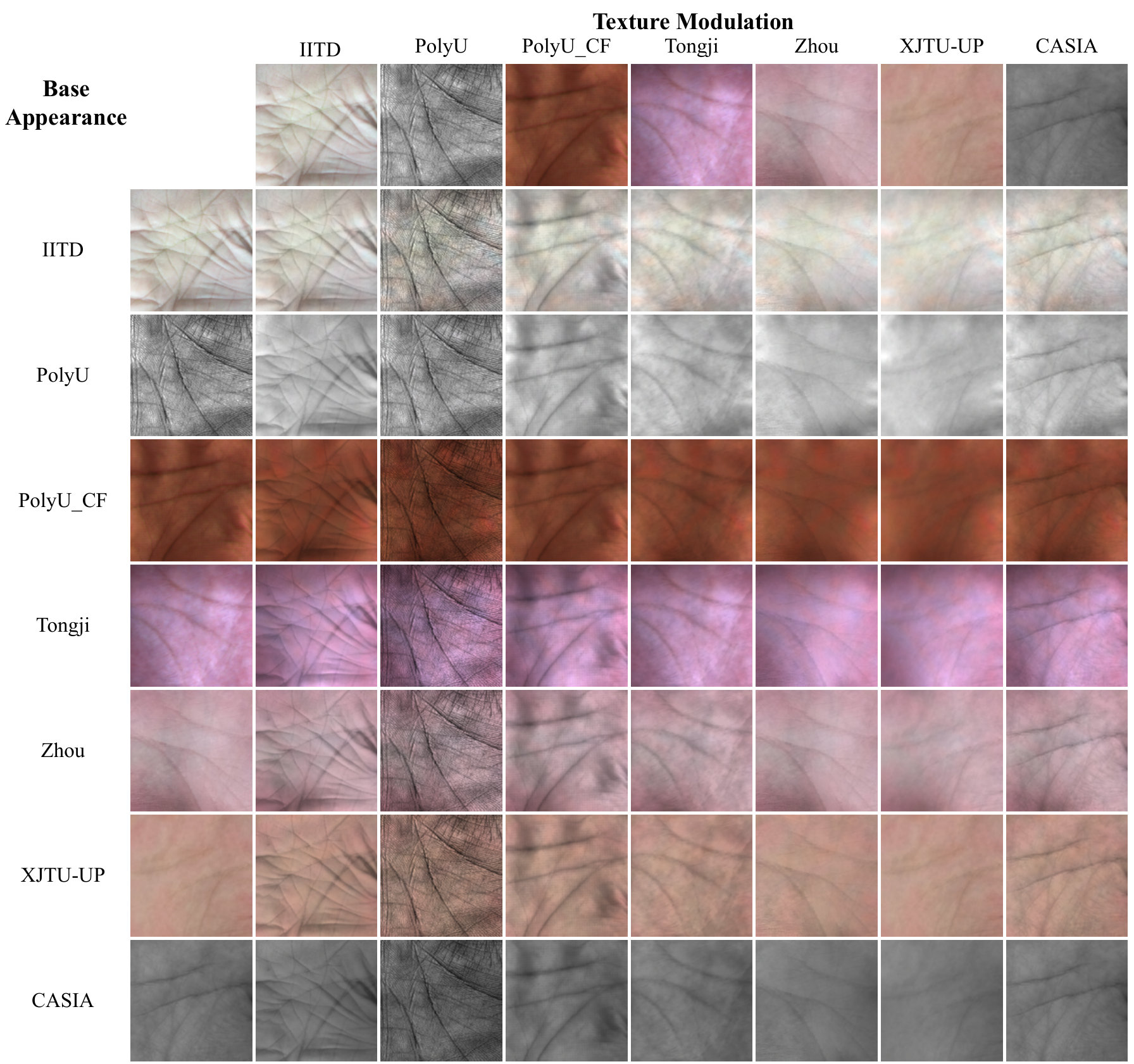}
    \caption{Cross-dataset factor exchange. Left-column images supply base appearance and top-row images supply texture modulation. Each interior image combines the corresponding factors through the fixed composition model. Diagonal entries show self-reconstructions.}
    \label{fig:exchange_matrix}
\end{figure}

Figure~\ref{fig:independent_factor_editing} illustrates independent editing of the four factors. We further evaluate cross-dataset factor exchange by combining base appearance from one image with texture modulation from another. For a base appearance donor $i$ and a texture donor $j$, we combine $(\hat C_i,\hat L_i)$ with $(\hat S_j,\hat a_j)$ through the fixed composition model:
\[
    \bar I_{i,j}=\hat C_i(1+\hat L_i)(1-\hat a_j\hat S_j).
\]
In Figure~\ref{fig:exchange_matrix}, each column holds texture modulation fixed while varying base appearance; each row holds base appearance fixed while varying texture modulation.

\paragraph{Exchange Protocol.}
For each ordered pair of distinct datasets, we assign one base appearance donor to each texture image by traversing a random permutation of the donor dataset, reshuffling after each full pass. For XJTU-UP and CASIA, we use the images from all acquisition conditions included in the main evaluation. Following histogram-based appearance evaluation~\cite{gong2025salut}, we compute the mean RGB-channel histogram correlation between the exchanged image and its base appearance donor. Texture similarity is the MTCC score between the exchanged image and its texture donor.

\begin{table}[!htbp]
\centering
\caption{Dataset-wide factor exchange. Rows supply base appearance and columns supply texture modulation. Cells report mean RGB histogram correlation to the base appearance donor / MTCC texture similarity to the texture donor (both $\uparrow$). Diagonal entries are self-reconstructions.}
\label{tab:factor_exchange}
\setlength{\tabcolsep}{3pt}
\renewcommand{\arraystretch}{1.15}
\small
\newcommand{\exchangehead}[1]{\begin{tabular}[c]{@{}c@{}}#1\end{tabular}}
\begin{tabular*}{\columnwidth}{@{\extracolsep{\fill}}l*{7}{c}@{}}
\toprule
\exchangehead{$B$ / $T$} & \exchangehead{IITD} & \exchangehead{PolyU} & \exchangehead{PolyU\_CF} & \exchangehead{Tongji} & \exchangehead{Zhou} & \exchangehead{XJTU-\\UP} & \exchangehead{CASIA} \\
\midrule
IITD & 0.99/0.99 & 0.18/0.92 & 0.89/0.87 & 0.86/0.90 & 0.90/0.86 & 0.91/0.85 & 0.91/0.88 \\
PolyU & 0.25/0.82 & 0.99/0.99 & 0.31/0.82 & 0.34/0.85 & 0.25/0.79 & 0.16/0.77 & 0.24/0.82 \\
PolyU\_CF & 0.90/0.86 & 0.49/0.91 & 0.98/0.97 & 0.90/0.87 & 0.90/0.83 & 0.87/0.82 & 0.89/0.85 \\
Tongji & 0.92/0.88 & 0.49/0.91 & 0.94/0.85 & 0.99/0.98 & 0.91/0.83 & 0.85/0.83 & 0.91/0.86 \\
Zhou & 0.91/0.88 & 0.26/0.92 & 0.89/0.86 & 0.86/0.89 & 1.00/0.98 & 0.90/0.84 & 0.90/0.87 \\
XJTU-UP & 0.87/0.90 & 0.01/0.93 & 0.76/0.87 & 0.68/0.90 & 0.83/0.87 & 0.99/0.98 & 0.83/0.89 \\
CASIA & 0.93/0.89 & 0.31/0.92 & 0.91/0.87 & 0.89/0.89 & 0.92/0.86 & 0.91/0.86 & 1.00/0.99 \\
\bottomrule
\end{tabular*}
\end{table}

\paragraph{Results.}
Across the 42 directed dataset pairs in Table~\ref{tab:factor_exchange}, mean RGB histogram correlation and texture similarity are 0.71 and 0.87, respectively, with equal weight per pair (242,982 exchanged images). Self-reconstructions on the diagonal are excluded. Exchanges involving PolyU yield lower histogram correlations; XJTU-UP base appearance with PolyU texture gives 0.01 histogram correlation and 0.93 texture similarity. Strong PolyU crease contrast may shift the recomposed intensity distribution through texture-induced attenuation while preserving donor texture.

\FloatBarrier
\section{Instance-Wise Refinement Ablations}
\label{sec:sup_iwr}
We further examine the effects of refinement steps and the response-overlap constraint on reconstruction fidelity, factor separation, and discriminative texture preservation. Evaluation uses all 2,600 IITD images and 7,752 PolyU images, with texture verification performed on $T=aS$ using MTCC. The optimization settings are given in Section~\ref{sec:sup_refinement}.

\begin{figure}[p]
    \centering
    \includegraphics[width=0.98\linewidth]{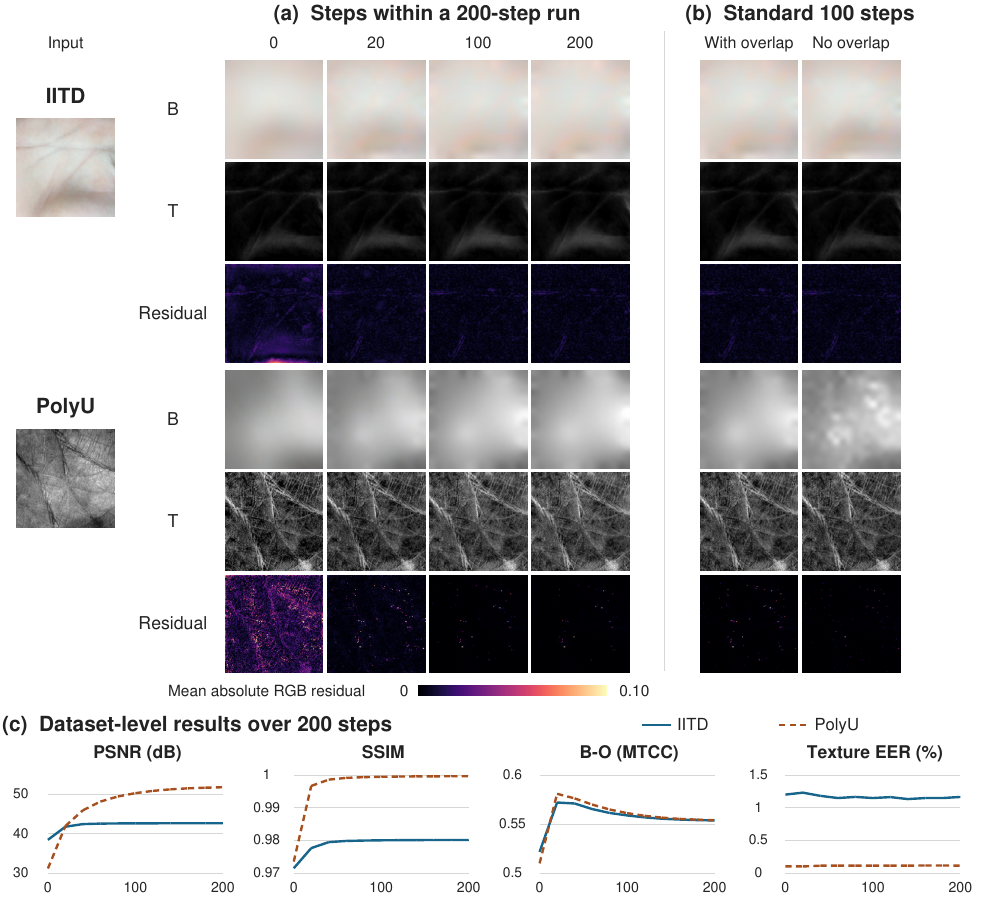}
    \caption{IWR on IITD and PolyU. Each input appears at left, followed by its base appearance $B$, texture modulation $T=aS$, and reconstruction residual. The residual is the per-pixel mean of $|\bar I-I|$ over RGB channels, shown on a common scale. (a) Four snapshots from a 200-step run. (b) The same inputs under the standard 100-step schedule with and without the overlap term. (c) Full-dataset metrics sampled every 20 updates along the runs in (a). Texture EER (\%) is computed using MTCC on $T=aS$ at each checkpoint.}
    \label{fig:iwr_comparison}
\end{figure}

\paragraph{Refinement Steps.}
Figure~\ref{fig:iwr_comparison}(c) reports dataset-level metrics every 20 refinement steps over a 200-step run with the overlap constraint. Within the first 20 steps, PSNR increases from 38.45 to 41.80\,dB on IITD and from 31.13 to 41.98\,dB on PolyU, reaching 42.68 and 51.88\,dB at step 200. Reconstruction gains saturate earlier on IITD. Later updates reduce B--O with limited changes in texture EER.

\paragraph{Overlap Constraint.}
For the comparison in Figure~\ref{fig:iwr_comparison}(b), we use the standard 100-step results reported in Table~\ref{tab:multidataset_final} as the reference. Removing the overlap term raises PSNR from 42.62 to 42.74\,dB on IITD and from 48.38 to 54.12\,dB on PolyU. Base appearance also retains more input texture. B--O rises from 0.564 to 0.607 on IITD and from 0.569 to 0.608 on PolyU. On IITD, texture EER rises from 1.16\% to 1.29\%. In the PolyU example, more of the palm pattern is visible in base appearance when refinement uses reconstruction loss alone. These results indicate that the overlap constraint reduces texture leakage into base appearance while maintaining high reconstruction fidelity.

\FloatBarrier

\end{document}